# Argus: Quality-Aware High-Throughput Text-to-Image Inference Serving System

Shubham Agarwal[*]
shubham3@berkeley.edu
UC Berkeley
Berkeley, CA, USA

Subrata Mitra
subrata.mitra@adobe.com
Adobe Research
Bangalore, India

Saud Iqbal[*]
saudiqbal886@gmail.com
Independent Researcher
Bangalore, India

## Abstract

Text-to-image (T2I) models have gained significant popularity. Most of these are diffusion models with unique computational characteristics, distinct from both traditional small-scale ML models and large language models. They are highly compute-bound and use an iterative denoising process to generate images, leading to very high inference time. This creates significant challenges in designing a high-throughput system. We find that a large fraction of prompts can be served using faster, approximated models. However, the approximation setting must be carefully calibrated for each prompt to avoid quality degradation. Designing a high-throughput system that assigns each prompt to an appropriate model and compatible approximation setting remains challenging.

We present Argus[1], a high-throughput T2I inference system that selects the right level of approximation for each prompt to maintain quality while meeting throughput targets on a fixed-size cluster. Argus intelligently switches between approximation strategies to satisfy both throughput and quality requirements. Overall, Argus achieves **10x** fewer latency service-level objective (SLO) violations, **10%** higher average quality, and **40%** higher throughput compared to baselines on two real-world workload traces.

## 1 Introduction

Text-to-image (T2I) services like Midjourney [14], Adobe Firefly [10], and OpenAI DALL-E [12] are widely used, with several cloud providers also offering T2I services [9]. Diffusion models (DMs) are the backbone of most image generation systems, with the Stable Diffusion (SD) [8] family being one of the most popular open-source implementations, widely available on platforms like HuggingFace [6]. Just as traditional ML workflows have seen a surge of interest in high-throughput inference systems [23, 37, 65], the rapid growth in T2I usage has created a pressing need for scalable, high-throughput T2I inference systems. However, T2I inference systems face unique challenges in delivering cost-effective, high-quality service under fluctuating real-world production workloads.

[*]Work done while working at Adobe Research.
[1]In Greek mythology, Argus was known for his exceptional vigilance.

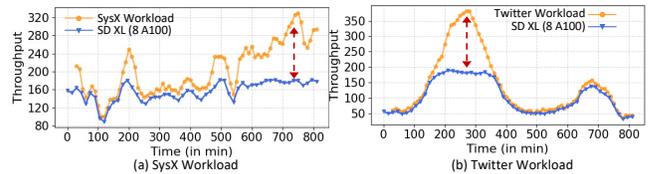

*Figure 1: Throughput fails to meet the peak load in production traces from* System-X, *a T2I service, and Twitter [3].*

**Challenges in meeting throughput:** Although large DMs are known for generating high-quality images, their inference latency is very high due to the iterative generation process requiring 50 to 100 denoising steps [4]. For instance, the large SD-XL model from HuggingFace [63] can take up to 10 seconds on an A10G GPU on AWS [7], and even 6 seconds on an A100-80GiB [1] GPU, to generate a single image. In Fig. 1, we show real workload patterns (throughput normalized to match the peak of a public trace, due to proprietary constraints) from a production system (System-X, where we sampled a window where request volume was close to the monthly median, to highlight the variability) (left) and a publicly available Twitter trace [3] (right), which has been used in prior ML inference evaluations [23, 65]. These demonstrate that a T2I inference system running SD-XL models on 8 A100 GPUs cannot reliably meet user demand during load spikes. Conversely, provisioning for peak load leads to significant resource wastage.

Batching has been widely used in prior works [32] and by ML practitioners to increase throughput for traditional (smaller) ML models such as ImageNet and YOLO, which are typically memory-bound. It is also heavily employed in state-of-the-art LLM inference systems [82], where the *decode* phase is memory-bound. However, as we observe in § 3.2, batching is ineffective in DM-based T2I systems because these models are compute-bound, and latency increases significantly with batch size, even from the smallest batches.

**Strategies to increase throughput:** We make several key observations (detailed in § 3.2) that guided the design of Argus.

First, while high-quality images are crucial for user satisfaction, many production prompts are *approximation-tolerant.* That is, when paired with faster, smaller, or more approximated models, these prompts produce images with indistinguishable quality differences. We confirmed this through quality metrics and a user study on popular T2I models (**Observation-1**). In § 2, we describe two prominent approximation strategies compatible with T2I: (a) using multiple model variants of different sizes (smaller means faster), and (b) dynamically accelerating a single model by reducing its computation through retrieval from a cache (approximate caching or AC).

| Baseline | Model Selection | Query Specific Approximation | Scaling | Use of Batching | Meets Threshold | Target Models |
|---|---|---|---|---|---|---|
| CLIPPER[32] | No | No | Auto | Yes | No | Disc. ML |
| Sommelier[38] | Yes | No | Fixed | No | No | Disc. ML |
| NIRVANA [20] | No | Yes | Single | No | No | T2I |
| TABI [74] | No | Yes | Single | No | No | LLM |
| Cocktail [37] | Yes | No | Auto | No | Best effort | Disc. ML |
| INFaaS[65] | Yes | No | Auto | Yes | Best effort | Disc. ML |
| Proteus[23] | Yes | No | Fixed | Yes | Yes | Disc. ML |
| Argus (ours) | Yes | Yes | Fixed | Conditional | Yes | T2I |

Table 1: Comparison of inference serving systems on various aspects.

Second, while using the largest (and slowest) model is overkill for many prompts, there is an optimal approximation level for each prompt. Prior works [23, 65] for small/discriminative models switch to less accurate (faster) models under high load. However, critically, these approaches assume model accuracy is uniform across inputs, making them input-agnostic. Using a model more approximate than what a prompt can tolerate leads to quality degradation. Hence, randomly assigning prompts to models with varying approximation levels proves to be significantly suboptimal (**Observation-2**).

Third, ideal prompt allocation is not always feasible, as workload patterns and prompt characteristics can independently vary. Thus, a good T2I inference system should find the best possible match between prompts and approximation levels to meet throughput with minimal quality loss (**Observation-3**).

Fourth, there is no silver bullet when it comes to finding a universal approximation strategy. Workload patterns and system states determine the optimal approach among the two primary techniques described. Switching among multiple models can incur significant overhead for large DMs in dynamic workloads, while approximate caching's retrieval overhead, usually minimal, can increase under network congestion or storage degradation (**Observation-4**). All these observations together motivate the key design principles that Argus embodies to thrive under real-world T2I workloads.

**Our system Argus in a nutshell:** Argus manages an inference-serving cluster with a fixed number of GPUs. It uses a combination of approximation strategies and quality-aware scheduling algorithms to consistently meet throughput targets with minimal degradation in output quality. Based on the system load, Argus solves an optimization problem to decide how many instances of each model should run at different approximation levels. It then determines what fraction of the input load should be assigned to each level to satisfy throughput constraints. Furthermore, at a micro-level, it employs a novel algorithm to match individual prompts to the most suitable approximation level. This ensures that the macro-level allocation is met, while also achieving the best possible prompt-level matching to improve generation quality at a micro-level.

**Argus vs. Prior Work:** Table 1 compares Argus with prior inference-serving systems. Sommelier [38], Cocktail [37], INFaaS [65], and Proteus [23] use multiple small models for discriminative tasks but do not make input- or query-specific approximation decisions. Proteus focuses on throughput, whereas INFaaS treats accuracy as a constraint and throughput as best-effort. Batching, effective for smaller or memory-bound models, benefits systems like CLIPPER [32], INFaaS, and Proteus, but not large-scale T2I models.

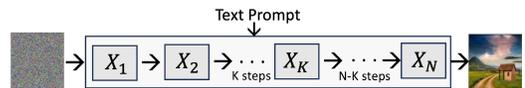

Figure 2: For a given prompt, a DM generates an image by iteratively denoising a Gaussian noise using the prompt as the condition.

INFaaS and Cocktail rely on horizontal autoscaling, which is inefficient for T2I inference due to cold starts, GPU scarcity, and low utilization. ARGUS, by contrast, runs on a fixed cluster and adapts approximation levels to meet throughput goals without on-demand scaling. INFaaS's horizontal scaling limitations for high-end GPU workloads are also detailed in Proteus [23], explaining why direct comparison with INFaaS and Cocktail is not meaningful.

Discriminative cascades like Tabi [74] and confidence-based routing methods like PICS [36] are also unsuitable: Tabi's token pruning and logit reuse do not apply to generative T2I models, and PICS's token-level confidence assumptions fail for high-dimensional outputs with iterative denoising (50–100 steps).

For evaluation, we compare Argus against the following baselines: CLIPPER, Sommelier, NIRVANA, and Proteus (details in § 5). **Key results:** Argus meets the throughput from both System-X workload and Twitter trace workload, under high load. It achieves 10% and 15% improvement over Proteus [23], 24% and 20% improvement over NIRVANA [20], and 5% and 7% over CLIPPER-HT [32], under high-load. Image quality improvements are 10% over Proteus and 15% over CLIPPER-HT. It also reduces SLO violations by 5-10× compared to both Proteus and NIRVANA around high points of System-X workload, which is also significantly fluctuating. Moreover, for a fixed-sized cluster configured for an average load of System-X, Argus provides high cluster utilization (90.82%).

**Contributions.** We summarize our key contributions as follows:
(1) We identify the unique practical challenges in scaling inference for large diffusion-based T2I models, including per-prompt quality tolerance and costly model-switch overheads, and explain why traditional batching and autoscaling are ineffective.
(2) We introduce *Prompt-Aware Scheduling*, a novel abstraction that leverages a lightweight classifier to predict each prompt's acceptable approximation level and an ILP-based solver to optimize prompt-to-approximation assignments in real time.
(3) Building on these, we design Argus, a high-throughput T2I inference system that dynamically switches among multiple approximation strategies to maximize quality and meet throughput under varying load and system conditions, as validated on two real-world production traces.

## 2 Background and Challenges

We provide background on text-to-image (T2I) generation, challenges in creating a high-throughput T2I serving system, and different approximation techniques.

**T2I with Diffusion Models (DMs):** DMs generate images by progressively denoising random Gaussian noise conditioned on text inputs [44]. During inference, these models iteratively refine the noise using the input text as a guide, resulting in a coherent final image (Fig. 2). Here, $N$ is the fixed number of diffusion steps used to denoise the image from Gaussian noise (e.g., N = 50 for SD models). While widely used, this diffusion process is computationally intensive and slow. Notably, diffusion modules are also employed in the final decoding stages of several recent autoregressive image generation pipelines [54], highlighting their continued importance across generative paradigms.

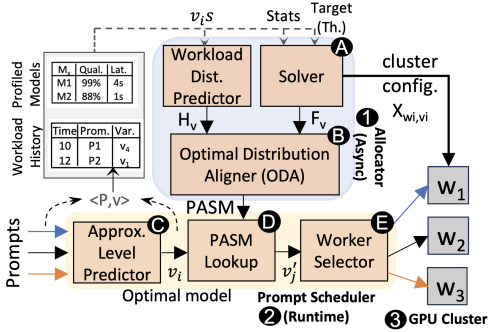
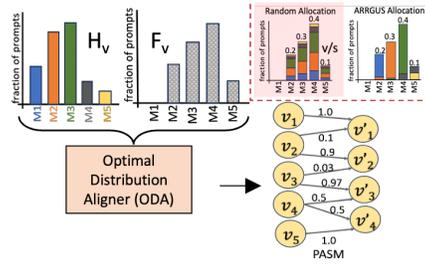
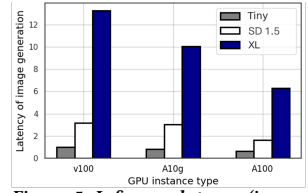

Figure 3: Argus design overview: It comprises 3 main parts: Allocator, Prompt Scheduler, and GPU Cluster.

Figure 4: Optimal Distribution Aligner (ODA) aligns the optimal model distribution $H_v$ and the load distribution $F_v$ to provide the Probabilistic Approximation Shift Graph (PASM).

Figure 5: Inference latency (in seconds) for Tiny, SD 1.5, and SD-XL models from Hugging Face [47] across three AWS GPU architectures. While SD-XL delivers significantly higher image quality, it incurs substantially higher inference latency.

**Image quality:** As gathering human preferences for generated images is not scalable, several automatic evaluation techniques have been proposed [25, 26, 29, 30, 39, 43, 50, 57, 72]. In this paper, we use PickScore [50], a state-of-the-art metric for assessing T2I alignment, and also leverage it during classifier training to supervise variant selection. PickScore uses CLIP's [43] dual-encoder to embed text and images into a shared representation space. Trained on the Pick-a-Pic dataset [50], which contains real user preferences over prompt–image pairs, higher PickScore values indicate stronger alignment between the generated image and the input prompt. Note that in this paper, we also evaluate image quality using human preferences from a user study, in addition to PickScore, to allow a more thorough validation of the output quality.

**High inference latency:** T2I models have high inference latency due to iterative denoising. Despite advances in GPU capabilities, newer and larger models like SD-XL still exhibit significant latency even on high-end GPUs. For instance, Fig. 5 shows the latency of SD-1.5 (2022) and SD-XL (2023) on NVIDIA GPUs V100, A10G [7], and A100, indicating that while older models run faster on newer GPUs, the latest models still incur significantly high latency.

**Batching is Ineffective:** Batching is commonly used in serving systems [32] to boost throughput by increasing GPU utilization without significantly raising inference latency [24]. However, as discussed in § 3.2, batching is practically ineffective for T2I inference serving, since latency rises sharply with batch size.

### 2.1 Approximations for Reducing Latency

**Faster model variants:** To reduce latency, a large and slow model (a.k.a. teacher) can be approximated with faster (and typically smaller) model variants (a.k.a. students) through distillation [60]. Distillation involves training a student model with fewer parameters and/or fewer denoising steps (temporal distillation [66]) to enable faster inference. Orthogonal techniques such as pruning [84] and quantization [42, 68] can also compress the original DM into smaller and faster variants. The conventional belief is that smaller or faster models yield lower image quality compared to larger models.

**Approximate caching (AC):** Recently proposed AC techniques can reduce denoising steps by reusing intermediate noise states cached during prior prompt processing [20, 86]. Specifically, AC retrieves an intermediate state from the $K^{th}$ iteration (out of $N$ total diffusion steps) of a previous generation. This state is then reconditioned with the new query prompt $\mathcal{P}_Q$ for the remaining $N-K$ steps, reducing latency by a factor of $(N-K)/N$. Intermediate states are stored in a file system and indexed using a Vector Database that holds prompt embedding vectors. Using similarity search, the most similar cached prompt is retrieved. Based on prompt similarity, an appropriate approximation level ($K$) is selected, and the corresponding noise state is fetched. As with smaller models, larger $K$ values yield greater speed-ups at the cost of lower image quality.

In Argus, we leverage both distilled (faster) model variants and approximate caching to sustain high throughput and image quality under varying conditions (see §4.6 for details).

## 3 Argus Overview

We now present an overview of our system, Argus, which aims to sustain high-throughput text-to-image (T2I) generation under load while minimizing quality degradation and latency SLO violations. We begin with two guiding concepts that inform our system design.

**Optimal Quality Image:** Let there be $N$ text-to-image models $\{M_1, ..., M_N\}$. For a given prompt $P$, let $\{q_1, ..., q_N\}$ denote the quality scores (e.g., PickScore [50]) of images generated by these models. We define an image as *optimal quality* image if its score $q_i$ satisfies $q_i > \delta \times max\{q_1, ..., q_N\}$. We use $\delta = 0.9$, consistent with prior work [20], meaning any image whose PickScore is within 90% of the value of the best quality image for that prompt can be considered to be of *optimal quality*.

**Optimal Model:** Among all models, the one that generates an *optimal quality* image with the lowest inference time is termed the *optimal model* for that prompt. We also say that the particular prompt has an *affinity* for this model, and refer to this as the prompt's **optimal model choice**.

### 3.1 High-Level Architecture

At a high level, Argus manages a fixed-sized GPU cluster and judiciously uses various approximation strategies (§2) and corresponding levels of approximation for T2I generation using diffusion models to meet the required throughput with minimal degradation in quality. For an incoming prompt, Argus first determines the

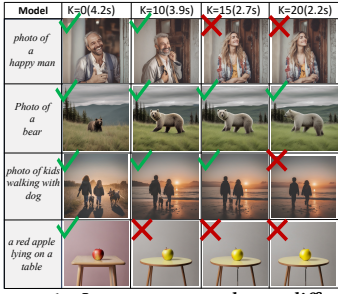

Figure 6: Images generated at different approximation-caching levels for AC.

| Model | Size | PyTorch | Accelerate | Latency |
|---|---|---|---|---|
| SDXL | 5.14s | 45.78s | 9.42s | 4.2s |
| SD1.5 | 3.44s | 19.90s | 5.56s | 3.84s |
| Small | 2.32s | 14.05s | 4.86s | 2.75s |
| Tiny | 0.63s | 11.78s | 2.91s | 2.18s |

Table 2: Different model loading times (seconds) and sizes (GBs) for DMs on A100.

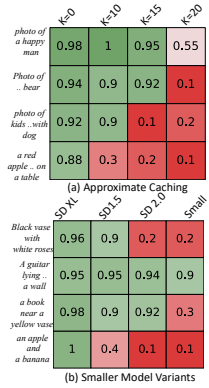

Figure 7: Votes from users on images generated by different approximation levels.

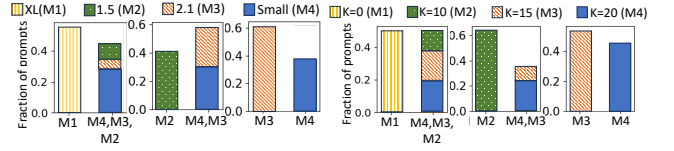

(a) Smaller Model variants     (b) Approximate-Caching

Figure 8: Distribution of prompts among their respective optimal model choices for a different model and AC variants. This figure highlights that both smaller models and faster AC variants can produce images of comparable quality to larger (or slower) models for a substantial fraction of prompts.

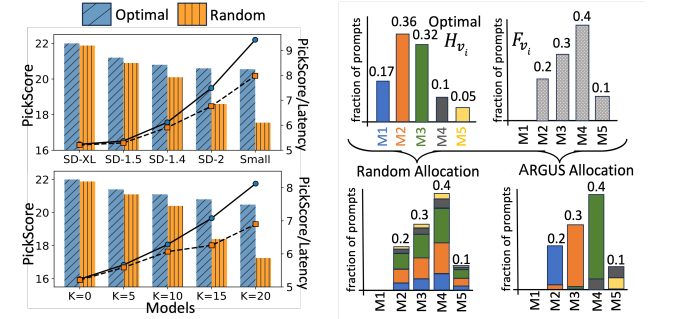

Figure 9: Average PickScores of images generated for prompts using their optimal model vs random assignment across all models. The top plot is for smaller model variants, and the bottom is for approximate-caching. We also show the PickScore-per-Latency.

Figure 10: For a target workload throughput, ARGUS computes the $H_{v_i}$ and $F_{v_i}$. Then, we see how ARGUS provides an optimal allocation strategy by carefully redistributing the prompts to other models as compared to random redistribution.

*optimal model* running at a particular approximate level using a classifier (Ⓒ in Fig. 3 and § 4.1). As the cluster may not be capable of handling all the prompts by its corresponding *optimal model* due to constraints on cluster size and throughput requirements, ARGUS probabilistically redistributes prompts to other models running on the cluster with different approximation levels (Ⓓ in Fig. 3 and § 4.3). For this, it looks up a *Probabilistic Approximation Shift Map (PASM)* to find the probabilities with which it can re-distribute the incoming prompts to new target models running at different approximation levels. The prompt is then routed by the Worker-Selector (Ⓔ in Fig. 3 and § 4.4) to an appropriate Worker hosting that model where the final image is generated. Periodically, ARGUS calibrates the cluster by recalculating (Ⓐ in Fig. 3 and § 4.2) how many workers should be used to host models at what approximation levels to meet the current load and also recalculates (Ⓑ in Fig. 3 and § 4.3) the redistribution probabilities for the PASM so that it can meet the throughput requirements with minimal quality degradation. ARGUS also switches (§ 4.6) between two different approximation strategies, selecting the one with the lowest overhead and quality degradation for the given workload and system state.

## 3.2 Observations Influencing ARGUS's Design

We elaborate on key observations from analyzing 10,000 real prompts from DiffusionDB [76], which influence ARGUS's design.

**Observation 1: Many T2I prompts are *approximation tolerant* – produce images of indistinguishable quality when a smaller or more approximated model is used.** Fig. 6 presents images generated at various approximation levels (skipping $K = [0, 10, 15, 20]$ steps) using AC. As shown, Prompt 2 can be adequately handled with higher approximation ($K = 20$ for AC), whereas Prompt 3 can tolerate less approximation, only up to $K = 15$ as the "dog" in the prompt disappears at a higher approximation level ($K = 20$). Factors such as prompt complexity and DM training design may influence this. To validate this, we collect user votes from 200 participants in Fig. 7(a) for AC. Similar findings are observed for smaller model variants in Fig. 7(b) based on user Yes/No votes.

To further validate this behavior, we analyze 10k prompts from DiffusionDB [76]. We consider four model variants: M1, M2, M3, and M4, where M1 (SD-XL) is slowest/largest and M4 (SD-Small) is fastest/smallest. We measure quality using PickScore and determine the optimal model for each prompt.

Let us denote the image quality from each of the 4 models as $q_1, q_2, q_3, q_4$ for a given prompt, where we measure quality using PickScore [50]. We designate model M1 as the **optimal model choice** if, for a given prompt, it is the fastest model capable of generating an *optimal quality image*. For prompts where M1 is not the optimal choice, we calculate the instances where $q_4 \geq 0.9 q_1$, $q_3 \geq 0.9 q_1$ (but $q_4$ is worse), and $q_2 \geq 0.9 q_1$ (but $q_4$ and $q_3$ are worse) to determine the *optimal model* choice.

In Fig. 8a, the left plot shows prompts optimally served by M1 (yellow bar). The stacked bars represent fractions served by M4, M3, and M2. The middle and right plots show analysis after eliminating M1 and both M1+M2, respectively. Fig. 8b shows a similar analysis using AC with four K values, yielding similar observations.

**Takeaways**: This observation motivated us to take advantage of various approximation strategies in ARGUS's design, to improve throughput without degrading quality.

**Observation 2: Indiscriminate approximations can significantly degrade quality, but when judiciously chosen for each prompt, quality degradation is reduced.** We observe that carefully allocating prompts to their *optimal-model* vs. indiscriminately allocating prompts to a model significantly improves image generation quality, especially for more approximated models. For example, as shown in Fig. 9 for production prompts, the average PickScore for images generated from the smaller SD-Small model is 17.4 when random allocation is used vs. 20.6 when only the prompts for which it is the *optimal model* are allocated to it. Similar behavior is present even for $K = 20$ with approximate-caching.

Therefore, previously proposed systems that used approximations for traditional/smaller ML models [23, 65] fail to maintain high-quality image generation under high load as they lack a prompt-aware approximation-level selection.

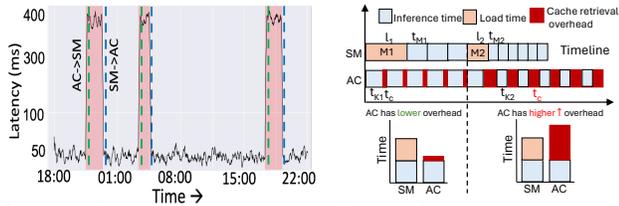

Figure 11: Showing an actual case on AWS when cache-retrieval latency spiked due to network issues prompting a switch from AC to SM.

Figure 12: SM faces model loading overhead (orange), while AC can face increased cache retrieval overhead (red) due to network.

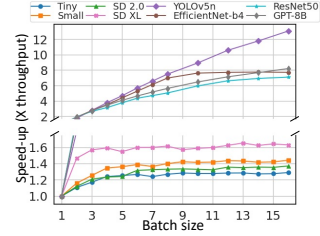

Figure 14: Speed-up in throughput vs. batch size on A100.

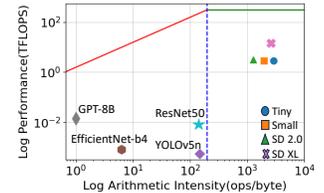

Figure 15: Roofline model for DMs and other DL models on A100 shows that DMs are highly compute-bound.

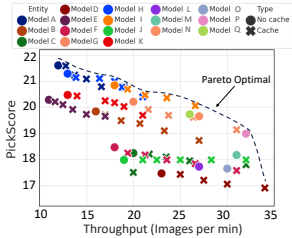

Figure 13: Pareto optimal plot shows AC offers better quality for a given throughput. A: SD-XL, H: SD-1.5, K: SD-1.4, D: SD-2.1, I: Small, N: Tiny, with X as their AC variants.

Table 3: UNet and VAE Encoder are the compute bottlenecks. Since UNet is used iteratively for N iterations, it remains compute bound for part of the total image generation time.

| Model | Component | #Param (B) | Model Size (GiB) | FLOPs (B) | Arithmetic Intensity |
|---|---|---|---|---|---|
| Tiny | Text Encoder | 0.123 | 0.229 | 7.208 | 29.287 |
| Tiny | UNet | 0.323 | 0.602 | 409.334 | 632.890 |
| Tiny | VAE Decoder | 0.050 | 0.092 | 2481.078 | 25066.363 |
| Small | Text Encoder | 0.123 | 0.229 | 7.208 | 29.287 |
| Small | UNet | 0.579 | 1.079 | 446.639 | 385.442 |
| Small | VAE Decoder | 0.050 | 0.092 | 2481.078 | 25066.363 |
| SD 2.0 | Text Encoder | 0.123 | 0.634 | 24.482 | 35.962 |
| SD 2.0 | UNet | 0.866 | 1.613 | 676.668 | 390.726 |
| SD 2.0 | VAE Decoder | 0.050 | 0.092 | 2481.078 | 25066.363 |
| SD XL | Text Encoder | 0.123 | 0.229 | 7.208 | 29.287 |
| SD XL | UNet | 2.567 | 4.782 | 11958.197 | 2328.574 |
| SD XL | VAE Decoder | 0.050 | 0.092 | 2481.078 | 25066.363 |

**Takeaways**: This motivated us to design a *classifier* for the corresponding approximation strategy in Argus. This classifier (described in § 4.1) can identify the *optimal model*.

**Observation 3: Infeasibility of ideal allocation.**

While allocating prompts to their *optimal model* produces high-quality images with minimal latency, it is not always feasible. To meet target throughput, Argus must find cluster configurations (using Ⓐ in Fig. 3 and described in § 4.2) where *optimal models* for certain prompts are unavailable as they might be too slow. It is also possible that to meet the incoming load, the distribution of prompts that must be allocated across models running at different approximation settings (let us denote this distribution as $F$) is different from the ideal distribution ($H$) of prompts across these models based on their *affinity* to the corresponding *optimal model*.

This situation is illustrated in Fig. 10 with real production workload data. Random redistribution from $H$ to $F$ leads to suboptimal quality (PickScore drops from 20.9 to 17.8) as approximation-sensitive prompts run at higher approximation settings than ideal.

**Takeaways**: When perfect approximation-setting is infeasible, prompts must be shifted carefully to alternate models to limit quality degradation. The optimal allocation strategy used by Argus (in § 4.3) achieves PickScore 19.5, minimizing quality loss.

**Observation 4: Between approximation strategies: approximate caching (AC) and using smaller/faster models (SM)—there is no single winner under all circumstances.**

**Approximation overhead:** Real production workloads show huge variations over time. To serve text-to-image inference at high throughput with minimal quality degradation, the serving system must switch between different approximation levels without introducing significant overhead. Diffusion models are large, and loading them from disk incurs significant overhead (Table 2). Due to GPU memory limitations, all models cannot be kept loaded simultaneously. Switching approximation levels by loading new models contributes to significant runtime delays. Fig. 12 illustrates this: in AC, higher values of $K$ accelerate inference without switching models, while in SM, switching between models introduces loading overhead (orange blocks). Aggregated view shows model-loading overhead (orange) is high in normal scenarios, favoring AC. Retrieval overhead (red) increases under network congestion, favoring SM.

In contrast, AC loads the largest model once and shifts approximation levels by adjusting denoising steps skipped ($K$), avoiding model-switching entirely. However, AC depends on external components—a vector database and storage backend—for cache retrieval. Under network congestion, AC may suffer from high retrieval overheads (Fig. 11), degrading throughput at high request rates.

**Approximation quality:** While both AC and SM can reduce inference latency, their quality-throughput tradeoffs differ. We compare 17 T2I models (A–Q) and apply AC with varying skip steps ($K$). In Fig.13, each model is shown as a colored point, and corresponding AC variants are shown as five black "X" markers with different $K$ values ($K$ = 5, 10, 15, 20, 25). The Y-axis denotes median PickScore over 10,000 real-world prompts from DiffusionDB[76], while the X-axis shows throughput in images processed per second per model instance. We observe that AC variants frequently lie on the Pareto frontier, indicating higher image quality at similar or better throughput than corresponding small or distilled models. While some distilled models offer good quality-latency tradeoffs, they are not always optimal across all conditions.

**Takeaways:** Our system, Argus, supports both approximation strategies and allows adaptive use of AC and SM. By default, Argus prefers AC due to its lower switching overhead and better approximation quality. However, when network conditions deteriorate or cache retrieval slows, it adaptively switches to SM (including distilled models) to ensure robust performance (see Section 5.6) by eliminating reliance on the external components required for AC. This fallback comes at the cost of model-switching overhead.

**Observation 5: Batching does not increase throughput for T2I generation as models are compute bound.**

In § 2, we highlighted that batching does not improve throughput for T2I model serving. T2I models are highly compute-bound. In Fig. 14, we compare speed-ups across batch sizes for non-DMs such as YOLOv5n [64], ResNet50 [41], EfficientNet-b4 [71], and GPT-8B [28], alongside various DMs. Notably, DMs show significantly slower speed-ups that plateau rapidly, highlighting their limited ability to benefit from batching. Over 90% of generation time is spent in the compute-bound UNet. Since latency is closely tied to latent spatial dimensions, the same trend holds for lower resolutions. While smaller models (e.g., SD-Tiny with 64×64×4) may show marginally better batching behavior, they still fall short of traditional DL models like YOLOv5, which can efficiently handle batch sizes of 16, whereas SD-Tiny hits bottlenecks around batch size 4.

In Table 3, we show FLOPs and *arithmetic intensity* of internal components in various DMs, and in Fig. 15, using a *roofline model* [48], we plot these alongside non-DM models to show that DMs are highly compute-bound (right side of the dotted-black line).

**Takeaways:** When batching is used, latency can be significantly higher. Argus avoids batching entirely due to its ineffectiveness in compute-bound scenarios. Discussions with engineers revealed that the same "no-batch" practice is also implemented in System-X deployment. Note that here "ineffective" refers to the inability to scale batch size in the same manner as traditional DL models.

## 4 Argus Design Details

### 4.1 Selecting *Optimal Model*

To find the *optimal model* for a prompt, Argus employs a Classifier (a.k.a. Approximation-Level Predictor Ⓒ in Fig. 3) that uses a BERT-based [49] architecture. It is trained in a supervised manner where, for each prompt, we generate images at different approximation levels and calculate their corresponding PickScores. The classifier then learns to predict, given a prompt, the model running at what approximation level is likely to be optimal. Separate classifiers are trained for each of the two approximation strategies: approximate caching (AC) and model variants of different sizes (SM).

Both classifiers are first trained offline using 10k samples from DiffusionDB [76], with minimal overhead. At initialization, when trained from scratch, the primary cold-start cost arises from image generation. For subsequent training sessions, however, Argus reuses images generated during normal operation, keeping the process lightweight. Retraining is triggered only upon significant data drift, detected when the median PickScore in the current window falls below the moving average of previous windows. This process runs off the critical path, does not impact generation throughput, and takes only [3.35, 8.34, 16.59] seconds for [2000, 5000, 10000] prompts on an A100 80GB GPU. We show the losses and PickScore during the training of the classifier in Fig. 19.

Please note that Argus does not use a mixed-mode strategy combining both approximation methods due to the complexity of training a classifier for $P \times Q$ classes, for $P$ model sizes and $Q$ levels of K for AC, as it requires extensive data to train.

### 4.2 Distributing Load Across Models

The Allocator ❶ component in Argus asynchronously calculates how many models and at what approximation levels must be deployed on the cluster to meet an incoming load. Then, it calculates the PASM that is used to decide how the incoming prompts need to be allocated across these approximation levels to minimize quality degradation while meeting target system throughput.

**Solver:** For a given system load, a fixed-sized cluster, and a set of model approximation levels with different inference latencies, the Solver (Ⓐ in Fig. 3) finds out how many model instances of each approximate-levels ($v_i$) should be run and what fraction of the load (distribution denoted by $F(v_i)$) should be served by models running at each approximation-level $v_i$, to meet the target throughput. When the AC strategy is used (which is the *default* as discussed in § 3.2), these approximation levels correspond to different values of $K$. When Argus uses the other approximation strategy, where smaller model variants are used, then the approximation levels for this solver correspond to different model variants. Quantized variants, profiled for latency and generation quality (see Fig. 13), are also treated as valid approximation levels in this setting.

In § 4.6, we discuss how Argus switches between these two approximation strategies as and when needed.

Let $Q = \{q_1, ..q_{t-1}, q_t, q_{t+1}, ..q_T\}$ be the prompt queries arriving over time. We calculate the expected workload $W_t$ in terms of Query per Minute (QPM) at time $t$ using the past workload. For a given $W_t$, the following optimization formulation is solved using Integer Linear Program (ILP) to maximize the quality of generation $\mathbb{Q}$ while meeting the target workload $W_t$.

$$\text{Maximize } \mathbb{Q} = \sum_{v_i} Q_{v_i} \cdot F(v_i)$$

$$\text{subject to } F(v_i) = \sum_w x_{v_i,w} \cdot y_w \qquad \sum_w y_w = W_t \qquad (1)$$

$$y_w \leq \sum_{v_i} P_{th}(v_i) \cdot x_{v_i,w} \qquad \sum_{v_i} x_{v_i,w} \leq 1$$

In Eq. 1, $y_w$ denotes the serving throughput of worker $w$. We profile the relative quality ($Q$) and peak throughput ($P_{th}$) for each approximation level ($v_i$) offline and regularly solve an Integer Linear Program (ILP) to determine $x_{v_i,w} \in \{0, 1\}$ and $y_w$. The ILP solver's objective is to maximize the overall generation quality $Q$, where $Q_{v_i}$ is the profiled average quality of variant $v_i$ subject to throughput and assignment constraints using integer decision variables. Here, $x_{v_i,w} = 1$ indicates that the approximation level $v_i$ is running on worker $w$, and $y_w$ represents the number of prompts routed to that worker per minute. $y_w$ is in *query-per-minute (QPM)*. The constraint on $y_w$ ensures that the allocated $y_w$ load does not exceed the profiled average throughput $P_{th}(v_i)$ if worker $w$ is assigned to $v_i$. The constraint on $x_{v_i,w}$ ensures that each worker $w$ runs at most one approximation variant. However, note that multiple workers $w_i$s can still share the same variant $v_i$. Additionally, $F(v_i)$ is the distribution of queries assigned to each approximation level $v_i$.

**Workload Distribution Predictor:** It predicts the estimated distribution of incoming prompts' affinity for *optimal models*, generating an affinity histogram $H(v_i)$. This prediction uses the distribution of *optimal models* for historical prompts, provided by the classifier, over a look-back window of 1000 prompts for both AC and SM. Our evaluations show that with a look-back window beyond 1000, $H(v_i)$ can be predicted with very low error for real workloads.

### 4.3 Probabilistic Approximation Shift Map

Argus is prompt-aware. For an incoming prompt, the Classifier predicts the *optimal model* that uses just enough approximation level such that output quality does not degrade. However, the Solver orthogonally calculates how many models at what approximation levels should be used and how the load should be distributed across them by only considering the *average* output quality and inference speed of the models running at different approximation levels. As the system load fluctuates, the distribution of loads across models with different approximation levels ($v_i$) changes over time, and therefore it may not be feasible to schedule an incoming prompt on its *optimal model*. Probabilistic Approximation Shift Map (PASM) is a data structure in the form of a *bipartite graph* (Fig. 4) that captures how prompts for which $v_i$ was the optimal approximation-level (as per the Classifier), can be probabilistically redistributed to alternate models running at different approximation-levels $v'_j$ while minimizing any degradation in quality.

As shown in Fig. 4, for each left node in *PASM*, the outgoing edges to the right nodes denote how a prompt targeted for the left node must be probabilistically redistributed to other approximation levels to meet the load. This *PASM* is calculated periodically by the Allocator, asynchronously, using the *Optimized Distribution Aligner (ODA)* algorithm that we propose in this paper. PASM is used at runtime by Prompt Scheduler (❷ and § 4.4) to schedule prompts to models of appropriate approximation levels at runtime based on the already calculated $v_i \to v'_j$ redirection probabilities.

**Optimized Distribution Aligner (ODA):** Argus keeps track of the incoming prompts and what *optimal model* was preferred by the Classifier. The Workload Distribution Predictor (Ⓑ in Fig 3) aggregates this information as a preference distribution over *optimal models* at various approximation levels, denoted by $H(v_i)$. Since $H(v_i)$ and the actual load distribution $F(v_i)$ that Argus must use, calculated by the Solver, might differ, the ODA calculates the redistribution plan in a quality-aware manner such that output quality degradation in minimized and stores it as *PASM* (Ref. Fig. 4).

PASM, which is the final output from ODA, shifts queries either to a slower/better model running at $v'_j$ st $P_{th}(v'_j) < P_{th}(v_i)$ or to the *closest* possible faster/worse model running at $v'_j$ st $P_{th}(v'_j) > P_{th}(v_i)$. This is done while trying to minimize the overall quality degradation $\mathbb{D}_Q$ defined as:

$$\text{Minimize } \mathbb{D}_Q = \sum_i \sum_{j: P_{th}(v'_j) > P_{th}(v_i)} \mathbb{P}(v'_j, v_i) \cdot H(v_i) \cdot D(v'_j, v_i) \quad (2)$$

Eq. 2 provides the theoretical basis of Argus. Given an empirically profiled degradation $D(v'_j, v_i)$, it defines a well-posed objective whose minimizer yields the smallest achievable workload-level quality loss. The Optimized Distribution Aligner (ODA) provably attains this by shifting prompts to the nearest feasible variant $v'_j$, minimizing $\mathbb{D}_Q$ under the degradation model.

Eq. 1 maximizes quality via allocation, while Eq. 2 minimizes expected loss from redistribution between $F(v_i)$ and $H(v_i)$. Empirically, $D$ increases super-linearly with the model speed gap $P_{th}(v'_j) - P_{th}(v_i)$, so ODA shifts only to the closest feasible $v'_j$. Argus assumes no fixed degradation form; $D$ is an explicit input, and ODA minimizes total expected loss across redistributions.

We describe the ODA algorithm, which uses $H(v_i)$ and $F(v_i)$ to compute transition probabilities $\mathbb{P}(v'_j, v_i)$ between approximation levels and minimize Eq. 2 to produce the PASM.

Algorithm 1 iterates over approximation-levels $v_i$ in *reverse* order of speed (from **right-to-left**, that is from $v_i$ to $v_{i-1}$, such that $v_{i-1}$ is slower as compared to $v_i$) and compares the corresponding $v_i^{th}$ positions in $H_{v_i}$ and $F_{v_i}$. If $H(v_i)$ is greater than $F(v_i)$ (line 3), which means that there are more prompts which have $v_i$ as *optimal model* than what can be served by existing workers running model at approximation-level $v_i$. In this case, it shifts these prompts to the immediately left, that is at $v_{(i-1)}$ (line 4). Notably, this leads to no quality degradation. In the second case, we can have fewer

**Algorithm 1** ODA algorithm

1: Initialize $H_{\text{old}} \leftarrow H_{v_i}, \mathbb{P} \leftarrow \{\}$
2: **for** $v_i$ in Approximation-levels$\{v_n, v_{n-1}, \ldots, v_1\}$ **do**
3:    **if** $H(v_i) > F(v_i)$ **then**
4:       $\mathbb{P}(v_{(i-1)}|v_i) \leftarrow \frac{H(v_i) - F(v_i)}{H(v_i)}$
5:       $H(v_{(i-1)}) \leftarrow H(v_{(i-1)}) + H(v_i) - F(v_i)$
6:       $H(v_i) \leftarrow F(v_i)$
7:   **else**
8:      **while** $H(v_i) < F(v_i)$ **do**
9:         **for all** $m \in \{1, 2, \ldots\}$ **do**
10:           shift$_{(i,m)} \leftarrow \min(H(v_{(i-m)}), F(v_i) - H(v_i))$
11:           $\mathbb{P}(v_i|v_{(i-m)}) \leftarrow \frac{\text{shift}_{(i,m)}}{H_{\text{old}}(v_{(i-m)})}$
12:           $H(v_{(i-m)}) \leftarrow H(v_{(i-m)}) - \text{shift}_{(i,m)}$
13:           $H(v_i) \leftarrow H(v_i) + \text{shift}_{(i,m)}$
14:           $\mathbb{P}(v_{(i-m)}|v_{(i-m)}) \leftarrow \frac{H(v_{(i-m)})}{H_{\text{old}}(v_{(i-m)})}$
15:         **end for**
16:      **end while**
17:   **end if**
18: **end for**
19: **return** $\mathbb{P}$ {$\mathbb{P}$ serves as the PASM}

prompts than what can be served by the GPUs for a particular $v_i$(line 8). While this can look apparently fine, but note, that if certain $v_i$ has fewer prompts, that is $H(v_i) < F(v_i)$, it means some other $v_i$ has more than allocated, and we need to make room for those here, at the current $v_i$. Therefore, we bring as many prompts as required to fill the gap from the immediate left (i.e., $v_{i-1}$) in (line 9), defined as shift(i,m). At each step, the probability (line 4,10) is computed using a fraction of shift divided by the total at $v_i$. This process continues towards the left (i.e., $v_{i-m}$) until the gap is filled, and we keep on repeating the process for all other bars to the left. Finally, we compute all transition probabilities using the step probabilities obtained to get the PASM.

$$\mathbb{P}(v'_j|v_i) = \mathbb{P}(v'_j|v'_{j-1}) \ldots \mathbb{P}(v'_{j-n+1}|v'_{j-n}) \cdot \mathbb{P}(v'_{j-n}|v_i)$$

**Proof for Optimality:** We show that the proposed ODA algorithm minimizes the quality degradation, as described by the objective in Eq.2. We proceed with a proof by contradiction. (1) First, when for a $v_i$, $H(v_i) > F(v_i)$(line 3) is *True*, without loss of generality, let us select a $v'_j$ to the further left of the immediate-left block $v_{i-1}$ to shift from $v_i$, instead of shifting to $v_i$. This choice would not degrade quality but since $v_{i-1}$ is still left with a room (which otherwise would be filled by $v'_j$), other prompts from an approximation-level $v_k$ to the left of $v_{i-1}$ will be forced to move to $v_{i-1}$ leading to their degradation in quality. Therefore, this scenario is untenable. (2) Also, when $H(v_i) < F(v_i)$(line 8) is *True*, let us select a $v'_j$ to the further left of the immediate-left block $v_{i-1}$ to shift from $v'_j$. This choice would increase the gap $(v'_j - v_i)$. Now, since the degradation function $D$ increases *super-linearly* as $v'_j - v_i$ increases, the choice becomes suboptimal. This establishes the optimality of our algorithm. Note that the *Earth Mover's Distance*, commonly used for measuring distribution shifts, is inadequate here because it is symmetric; however, the distinct implications of left-to-right and right-to-left shift on quality require a more nuanced approach.

In generative T2I, quality–approximation tradeoffs remain empirical. However, the ILP and ODA formulations provide a theoretical sketch for optimality in workload-level allocations.

## 4.4 Prompt Scheduler

The `Prompt Scheduler` ❷ receives an input query prompt $P_Q$ and uses a trained classifier to predict the *optimal model* approximation level $v_i$ (§4.1). The appropriate classifier is selected depending on the active approximation strategy being used. Then, PASM (§4.3) is used to select the final model operating at approximation level $v'_j$ for the prompt. This selection ensures the current workload requirement is met with minimal quality degradation.

The `Worker-Selector` (❽ in Fig. 3) routes the prompts to one of the workers running at $v'_j$. If only one worker is operating at $v'_j$, the decision is straightforward. However, when multiple workers are available, Argus adopts a *greedy* approach, routing queries to minimize the total processing time. The optimal worker $w$ running at $v'_j$ is selected as:

$$w = \arg\min_w R_{queue,w} \times t_{\text{proc},v'_j} \quad (3)$$

where $t_{\text{proc}}$ is the image generation time for approximation-level $v'_j$ for the model operating at worker $w$, and $R_{queue,w}$ indicates the number of requests queued at worker $w$.

## 4.5 GPU Worker

Each `GPU worker` ❸ processes prompt requests to generate images based on resource allocation and scheduling decisions made by the system. To optimize for latency, Argus processes each prompt individually with a batch size of 1 (§3.2).

Each worker operates with either an approximate level $K$ for AC or one of the smaller model variants. In AC, the worker searches a Vector Database (VDB) for the closest previous prompt, retrieving the corresponding intermediate noise state from Amazon Elastic File Storage (EFS), for the current $K$ at which the model on the worker is functioning. This state serves as the initial noise for generating the output image through iterative denoising, involving $N - K$ steps. Note that AC requires network calls to both the VDB and EFS.

In contrast, for smaller model variants, the worker generates an image starting from random Gaussian noise. This strategy may require loading/unloading of the model variant on the worker when the approximation level changes. However, for AC, the same model is used, with only a $K$ switch.

## 4.6 Switching Approximation Strategies

Argus determines which of the two approximation strategies (approximate caching AC vs. smaller model variants SM) is better, depending on the load and network connectivity.

By default, Argus uses AC (as discussed in § 3.2) as in normal situations, overhead latency for cache retrieval is orders of magnitude less compared to the latency savings from reducing denoising iterations. However, Argus constantly monitors the cache retrieval latencies. If, due to network failure or congestion, the retrieval latency increases substantially, then using AC will add overhead to each prompt which cumulatively can degrade the performance of the cluster, and in the most serious situation, it can fail to process any prompts due to unavailability of the VDB or EFS, which is the cache-storage. In that case, Argus initiates a switch to SM approximate strategy. However, while it runs in the SM mode, it constantly also runs test retrieval requests in the background to check if the network situation has improved and then again initiates a switch to a more optimal AC strategy.

All the internal components and the workflow fundamentally remain identical across these two strategies. Just based on the current strategy, the `Classifier` is switched, the solver uses the latency-quality values corresponding to the approximation levels for the specific approximation strategy, and PASM is recalculated.

**AC ↔ SM Switch** Argus runs AC using SD-XL models on all workers by default. When cache retrieval latencies become overhead, Argus switches to using SM. During the switch, Argus first starts serving prompts using the loaded SD-XL models but without caching to eliminate downtime. Concurrently, it loads smaller models on all workers. This is possible because GPU memory (i.e., 80 GB HBM for A100 GPU) can hold two DM models (which can be ≈15 GB for SD-XL and SD-1.5) together. Note that using the largest model without cache leads to a drop in throughput. To cover for it, Argus starts using smaller model variants as soon as they are loaded. Since models of different sizes may load separately, Argus uses the solver described in §4.2 to solve optimization as soon as small model variant(s) finish loading to incorporate them into the system and start diverting the load to it. During the switch, the solver uses a 1.5× margin to divert more load to a smaller model to cover for the throughput drop. As the new model becomes operational, the previous model is unloaded in the background. Conversely, when switching from SM to AC, all large models, being identical, are loaded almost in parallel. This allows Argus to do a full transition from smaller models back to using AC, and the smaller models are then unloaded in the background. This strategy also avoids running in any mixed-mode configuration.

## 4.7 Implementation Details

Argus is deployed on a cluster of 8 NVIDIA A100 GPUs (80 GB HBM2e), with each GPU running one model variant inside Docker containers (`Workers`). All models are served using PyTorch with CUDA support. The `Allocator` and `Prompt Scheduler` execute on a shared CPU node, while ILP-based load assignment is solved every minute using Gurobi via `sklearn`. We use `accelerate` [2] to optimize model loading and device placement.

AC is enabled through a Qdrant Vector Database hosted on a 32-core AMD EPYC 7R32 CPU instance. All GPUs in the cluster connect to this shared VDB. For each incoming prompt, the classifier predicts a reuse step $K$, retrieves the nearest prompt embedding from VDB, and fetches the corresponding intermediate noise (144 KB) stored in AWS Elastic File System (EFS). The diffusion model resumes from step $K$ to complete the remaining $N-K$ steps ($N$ = 50). Each prompt stores intermediate states at $K$. During tail latency conditions, Argus selects smaller variants from $SM$ to satisfy SLO constraints. Baselines are implemented using Proteus [23].

## 5 Evaluation

We conduct a comprehensive evaluation of Argus to demonstrate the benefits of prompt-aware, quality-driven approximation for scaling text-to-image (T2I) serving. Our results (§5) show that by selecting the fastest model that preserves image quality for each prompt, Argus achieves up to 10× fewer SLO violations, 10% higher image quality, and up to 40% greater throughput compared to state-of-the-art baselines. These gains are supported by a human perception study with 186 participants (§5.4), where Argus-generated outputs are consistently preferred over baselines. Additionally, Argus maintains performance under increasing load and bursty demand, and rapidly recovers from hardware failures. We now describe the experimental setup, followed by results along four axes: system performance, load resilience, classifier impact, and runtime analysis.

## 5.1 Experimental Setup

**Base Model:** Argus uses Stable Diffusion XL (SD-XL) in AC mode with the *FP16* variant as the base model. This model comprises approximately 2.3 billion parameters [63] and operates in a 96-pixel latent space, executing $N = 50$ denoising steps to generate a final image of size $768 \times 768 \times 3$. On an A100 GPU, the average inference time is 4.2 seconds (close to the median).

**Approximation Strategies:** For different approximation levels, Argus uses 6 values of $K=[0,5,10,15,20,25]$ for AC and for SM employs 6 commonly used small/distilled DM variants *Tiny-SD, Small-SD, SD-1.4, SD-1.5, SD-2.0*, and *SD-XL* from HuggingFace [5].

**Test bed:** Our implementation of Argus is deployed on 8 NVIDIA A100 GPUs with 80 GB of GPU memory. Every minute, we solve to get an optimal solution for workload assignment (same as [23]). As VDB we use Qdrant [15] (version 1.3) instance running on a 32-core AMD EPYC 7R32 2.8 GHz CPU.

**Baselines:** We evaluate two versions of our system:
**Argus:** The full version with all components.
**PAC - Prompt-Agnostic-Argus:** This version *does not* use the ODA🅑 and Classifier🅒 (Ref. Fig. 3) and therefore randomly routes prompts to workers, same as [23]. However, it still uses the logic to determine the best approximation strategy between AC and SM, unlike [23], which only uses SM.

We compare against the following baselines:
**Proteus** [23] uses multiple small/distilled models (SM) and, based on the load, distributes traffic in a prompt-agnostic way, to meet throughput. We use the same 6 DM variants as used by Argus.
**Clipper** [32] is a static inferencing system that requires the user to decide what model to run. We have two variants:
- **Clipper-HA** (High Accuracy): Where we use the most accurate SD-XL model on all GPUs.
- **Clipper-HT** (High Throughput): Where we use the fastest Tiny-SD model on all GPUs.

**Sommelier** [38] solves the optimization at the GPU level rather than the cluster level. It assesses the workload for each GPU and selects an appropriate model (out of the 6 variants mentioned). It recommends alternative models based on the single-server workload, enabling a switch from high-accuracy to low-accuracy variants.
**NIRVANA** [20] paper described a single deployed instance of the largest model (SDXL model in our setup) with AC to control approximation-level ($K$) based on the prompt. It does not describe how to achieve high throughput under high load. We extend this for the cluster setting where we replicate Nirvana on each worker and distribute the load uniformly.

**Workloads:** We evaluate with a combination of production and synthetic workloads, aiming to capture both real-world and a variety of specific patterns. For realistic system loads measured as queries/second (QPS), similar to several prior works [23, 32, 37, 65], we use the Twitter traces collected over a month in 2018 to ensure that our evaluation includes the workloads used by the baselines. Moreover, recent work on inference serving [83] noted that this trace resembles other real and production inference workloads, showing both diurnal patterns and unexpected spikes [67]. We also use the QPS pattern from another proprietary real-world *text-to-image* service trace from *SysX*. For the *SysX* trace, we normalize it to the same min-max range as the Twitter trace to anonymize the real workload numbers. From a month-long trace, we chose a day whose request volume was close to the monthly median and used a contiguous 800-minute slice.

In addition, we created a bursty synthetic workload featuring interleaved periods of low and high query demand, generated through a Poisson process for query inter-arrivals to introduce macro-scale bursts. For stress-testing, we also generate a diagonal, increasing QPS pattern, where workload linearly increases from a rate of 50 QPM to over 600 QPM over a time of around 800 minutes.

We use prompts from *DiffusionDB* [76] where we maintain the prompt order based on the dataset's arrival sequence. This approach, along with the chosen parameter values, aligns with prior works and allows us to stress-test our system under a variety of realistic and synthetic workload conditions.

**Evaluation Metrics:** We compare Argus against the baselines using the following metrics: (i) *Throughput*: Number of queries served in QPM (queries per minute). (ii) *Effective Accuracy*: Accuracy measured using *PickScore*[50], averaged over all successfully served queries, i.e., queries completed within their latency SLO. (iii) *SLO Violation Ratio*: Fraction of queries that exceed the latency SLO, defined as 3× the latency of the largest model (SD XL)[23].

## 5.2 System Performance of Argus

We first show results for an end-to-end comparison of Argus with the baselines on three workloads in Fig. 16.

In the Twitter trace workload (left plot in Fig. 16), we see that Argus consistently offers the lowest drop in relative quality (except Clipper-HA), and stays above 90% at all times. Amongst all the scaling approaches, it offers the lowest SLO violation ratio (less than 5%) with the highest system throughput to meet the load. This comes from the fact that Argus minimizes the overheads of loading models as workload changes, as it uses AC variants by default with the same *SD XL*. Its superior quality is also attributed to the use of prompt-aware *optimal model* variant selection.

*Clipper-HA* provides the highest relative quality (≈100%) as it never swaps out high-accuracy models for low-accuracy models. However, it suffers from the highest number of SLO violations (25% in aggregate). On the other hand, *Clipper-HT* has significantly lower SLO violations (≈5%) and significantly higher(30%) throughput than *Clipper-HA*, though it comes at the cost of quality (at just 85%). Similar to *Clipper-HA*, NIRVANA also offers around 94% average quality and is prone to drop in quality (drops to just 5% at max). However, it suffers from relatively lower throughput and high SLO violations (20%) as it cannot adapt to an increase in workload by shifting the prompts to more approximate models in a controlled manner. The dynamic scaling baselines, *Sommelier* and *Proteus*, perform better in terms of throughput and SLO violations at stable workloads (between 500 and 700 minutes).

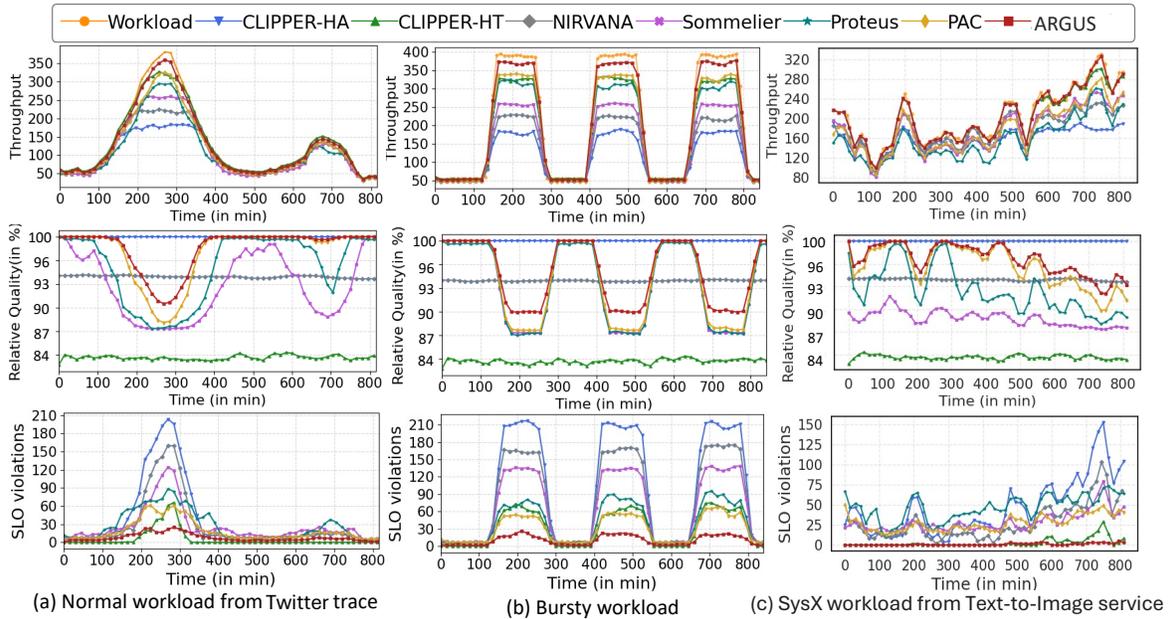

Figure 16: End-to-end performance comparison on Twitter, SysX, and synthetic workloads. Argus consistently meets incoming load with the lowest drop in quality and lowest SLO violation (up to 10x lower) compared to the baselines.

However, they suffer (25–30% SLO violations between 100 and 400 minutes) as workload changes due to high loading latencies and their limited ability to quickly switch to faster models. Also, their generation quality remains subpar (dropping below 90% relative quality at peak load), as they use small or distilled model variants with weaker quality-throughput tradeoffs and dispatch prompts in a prompt-agnostic fashion. Importantly, our comparison with *Proteus* also covers distilled-model baselines, as it uses the same set of distilled SM models for generation.

We repeat the experiment with a synthetic bursty workload. Here, we see that the aggregate throughput of Argus is highest with the least number of SLO violations (less than 5%). All other scaling baselines, *Sommelier* and *Proteus* suffer from high SLO violations and low throughput. We can see in Fig. 16 that as the load steps up, the baselines suffer from high SLO violations as they react slower due to large switching times (due to huge loading overheads). We also see that although NIRVANA generates high-quality images (at 94-95% relative quality), it suffers from a large number of SLO violations (up to 40-50%) at high load due to the inability to switch to a faster variant. These results demonstrate Argus 's ability to balance quality and throughput under abrupt load shifts.

Finally, we evaluate the system on another workload trace from a T2I production trace from *SysX*. We see Argus outperforms all scaling baselines by offering high throughput that almost always meets the loads and very low (less than 1%) SLO violations. Its generation quality is also best, apart from CLIPPER-HA, which suffers from low throughput and high SLO violations. Due to a jittery workload, Proteus and *Sommelier* suffer from high SLO violations and low throughput due to model switching overhead. PAC serves as a competitive baseline, similar to *Proteus*, but integrates both AC and SM. However, Argus 's Classifier + ODA significantly outperforms PAC, achieving 15% fewer SLO violations, 5% higher generation quality, and 12% greater throughput on the System-X workload by incorporating *prompt-aware* routing.

### 5.3 Stress-test under Increasing Workload

In Fig. 17, we show how Argus performs, compared to other systems, under a synthetic workload that keeps on increasing from a very low (40 QPM) to an extremely high load (540 QPM) for our test setup. We can see that at a very low load, as expected, all the systems perform nearly similarly in terms of throughput and SLO violations, as whatever queries come, get processed easily. For the relative quality of the generated image, all the systems except CLIPPER-HT and NIRVANA use the largest model or the most accurate mode ($k$=0 for Argus) of producing the image, leading to 100% quality. However, as the load increases, we see a diverging trend. Notably, Argus keeps up its throughput closely matching the increase in the load, it prevents quality degradation much better compared to the baselines (except Nirvana beyond 350 QPM), and keeps the SLO violations the lowest (except occasionally beaten by CLIPPER-HT). Recall, both Nirvana and CLIPPER-HT do not use accuracy-scaling. Nirvana suffers from bad throughput and SLO violation under high load, and CLIPPER-HT always provides a relatively poor-quality image as it only runs the smallest model. Under extremely high load (> 560 QPM), Argus has to use the highest level of approximation ($K$=25 for AC) on all 8 GPUs to sustain the load, and therefore quality drops. Beyond that, the throughput and quality saturate, and SLO violation keeps on increasing, showing the limits of accuracy-scaling for that particular cluster, which would require horizontal scaling (more GPUs) beyond this point.

### 5.4 Human Perception Study

Prior T2I studies compared models in isolation, not human perception under system-level adaptive approximations. We surveyed 186

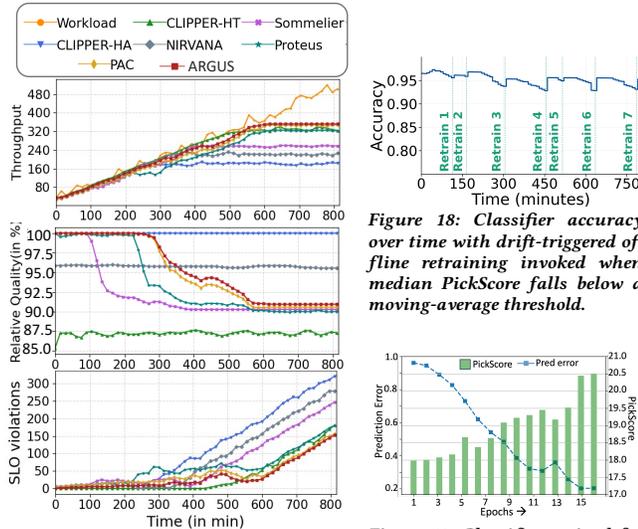

Figure 17: Performance under Increasing Workload. At high loads, ARGUS offers the highest throughput and lowest SLO violations with high generation quality.

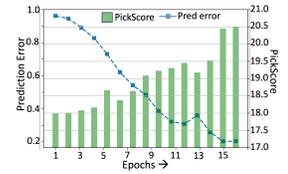

Figure 18: Classifier accuracy over time with drift-triggered offline retraining invoked when median PickScore falls below a moving-average threshold.

Figure 19: Classifier trained for more epochs has a low prediction error, which in turn enables better quality (higher PickScore).

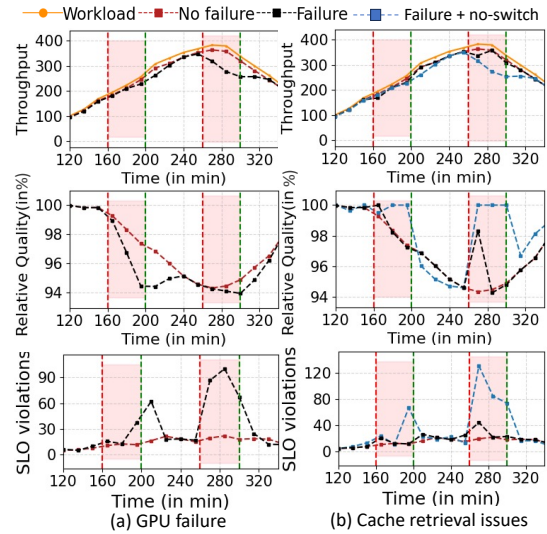

Figure 20: Impact of GPU failures on ARGUS performance under the Twitter workload. Failures trigger fallback to fast (K=25) variants, reducing image quality and increasing SLO violations by 3-5×. Performance also degrades when cache retrieval fails without adaptive switching.

participants (age 22-28; 42% female; 71% with graduate degree;) who rated images from ARGUS vs. baselines for prompt-relevance and overall-quality under load-conditioned scenarios. ARGUS achieved 82%/70% suitability ratings (prompt-relevance/overall-quality) vs. PAC (63%/46%), Proteus (59%/43%), and CLIPPER-HT (41%/35%). SD-XL scored 94%/89% but is unscalable. Results are statistically significant ($p < 0.05$), quantifying system-level effects that automated metrics miss. While this survey shows the surveyed users tend to prefer ARGUS, a comprehensive analysis of user bias across demographic is beyond the scope of this paper.

### 5.5 Generation Quality w.r.t. Classifier Accuracy

Random variant selection degrades quality relative to the classifier. In Approximate Caching (AC), PickScore falls from 20.8 (classifier) to 17.6 (random), a −15.4% drop; in Smaller Models (SM), it falls from 20.6 to 18.2, a −11.7% drop. We therefore use the classifier for routing in all subsequent results. Fig. 18 reports classifier accuracy over time with drift-triggered offline retraining (8B, 8 epochs per refresh). Retraining is invoked only when the median PickScore dips below a moving-average threshold; in an 800-min SysX workload, it triggered 7 times (≈3.5k new images per event) with no throughput impact. Accuracy can dip on highly out-of-distribution prompts but remains within targets for realistic T2I workloads. Fig. 19 examines how classifier quality translates to image quality in Argus: as training/validation loss decreases, average PickScore increases, for example, training loss 1.0 → 0.1 raises PickScore 18.0 → 20.6, underscoring the need for accurate prompt-aware routing.

### 5.6 Handling of System Faults

We simulate two possible failure/degradation scenarios.

*GPU failure*: We simulate worker failure where we switch off half, i.e., 4 out of the 8 GPUs, *twice* for around 40 minutes, once under relatively moderate load and another under high load, shown in *red background* in Fig. 20(a). Note, during the first failures, the effective throughput of the system rarely degrades as ARGUS's solver detects it within a minute, as it runs periodically and re-allocates the load on the remaining GPUs. But to accommodate this increased load per GPU, ARGUS now uses higher approximation levels (higher K), leading to quality degradation. In the second failure, the load is already so high that even with re-allocation, the throughput can not immediately recover, as existing GPUs are already running close to the highest approximation levels. Therefore, quality can not degrade any further to improve throughput.

*Cache Retrieval issues*: We also simulate a network failure scenario, where the connection to the VDB and cache storage fails. ARGUS detects this situation by monitoring cache retrieval latency and changes the approximation strategy to SM. During the AC →SM switch, it first goes for a fallback and runs large models on all GPUs with no AC (K=0). Since this is the slower version of the model, at high loads, we see in Fig. 20(b), the effective throughput is hurt, with higher SLO violations. Although quality improves momentarily, it is at the cost of slower inference serving. Then, as soon as smaller model variants become active, we see performance getting back. The violations are even higher in the second failure due to higher loads. In the plot, we also show a black line for handling failure without allowing a switch. In this case, performance is severely hit (only quality improves at the cost of performance) due to the use of all large models with no AC (K=0) at high loads.

### 5.7 Other Runtime Analysis and User Study

**High vs. Low Load Behavior:** ARGUS dynamically adapts to varying load conditions using prompt-aware routing. During light loads, it selects more accurate (lower approximation level); under peak load, it switches to less accurate (higher approximation level) variants to sustain throughput. This adaptation preserves image quality while improving average throughput and reducing SLO violations. While Twitter is mostly low-load, *Bursty* and *SysX* include high-load periods. Across all workloads, ARGUS consistently outperforms baselines in both throughput and SLO adherence.

**Variant Switching Overhead:** Argus minimizes variant switching, avoiding the frequent model loading overheads seen in Proteus and PAC, which switch in 27–42% of requests. This results in 15–20% higher throughput and fewer SLO violations.

**Cluster Utilization:** Argus achieves 1.5–2× higher GPU utilization compared to static over-provisioning for peak workloads across two real-world traces. Utilization improves from 37–60% under peak provisioning to 71–91% with Argus.

**Solver and Predictor Performance:** We evaluate the efficiency and accuracy of Argus 's core optimization components:

- **Solver Scalability:** The ILP solver scales well with cluster size, computing model placements in under 100 *ms* even for clusters with tens of GPUs.
- **Predictor Accuracy:** The workload distribution predictor estimates $H_{v_i}$ with high accuracy, achieving an L2 error of ~0.01 across workloads.

**User Study:** We conducted a user study with 186 participants to assess image quality under three load conditions across different serving strategies, using prompts from DiffusionDB. Argus outperformed all scalable baselines, with **82%** and **70%** of its images rated suitable for prompt relevance and overall quality at **mid** and **low** loads. PAC followed with 63% and 46%, aided by its prompt-aware design. Proteus achieved 59% and 43%, highlighting AC 's quality benefits. CLIPPER-HT, which always uses the smallest model, saw only 41% acceptance. SD-XL reached 94% but is not scalable.

## 6 Discussion

**Generalizability and Applicability Beyond Diffusion-Based T2I** Our Argus design, anchored by a prompt classifier and an ODA scheduler, naturally extends to any ensemble with monotonic accuracy–latency variants. While we demonstrate Argus on iterative diffusion models, where denoising steps ($K$) serve as the approximation knob, the same architecture applies to transformer-based generators or autoregressive decoders. In each case, one can redefine the approximation axis (e.g., early-exit thresholds or token-sampling strategies) and adjust the degradation penalty accordingly.

Notably, our current AC–SM switch leverages the negligible reload cost of adjusting $K$, whereas swapping in a smaller model incurs GPU loading latency. However, this choice is using a heuristic; future work could explicitly factor these load costs directly into the ILP (or prefetch warmed SM variants) to avoid transient SLO violations. This insight also directly maps to other models, such as those using early exits, which avoid reloads by operating within a single model, making Argus applicable across any model family that offers a quality–latency spectrum without altering the core scheduling logic. Considering load-time dimensions in the scheduler could further reduce overhead and improve efficiency, which is one of the novel design contributions of this work.

**Operational Boundaries and Performance Limits** Every approximation strategy eventually reaches its saturation point. When load forces all variants to their highest approximation levels (e.g., $K_{max}$ in caching or the smallest available model), image fidelity degrades beyond acceptable thresholds. This saturation serves as a clear signal for horizontal scaling. While Argus does not currently implement autoscaling, it can be readily extended to do so by using this saturation signal to trigger scaling decisions automatically.

Another limiting factor is classifier accuracy, which is equally critical: our ablation (§5.5) shows that even a 0.1 increase in loss yields a noticeable drop in the PickScore, defining a threshold below which quality-driven scheduling loses its benefit. In practice, sudden shifts in prompt characteristics can reduce cache hits and degrade classifier predictions. While Argus currently retrain the classifier periodically, a more robust design could leverage online or active learning to adapt in real time. We leave this as future work.

Finally, real-time adaptation relies on the performance of both the solver and predictor. Our ILP solver computes new assignments in under 100 ms for clusters of tens of GPUs, and the workload predictor maintains an error below 0.01 using a 1000-sample window. However, as cluster sizes grow or reconfiguration intervals shrink, solver latency could approach SLO budgets. In such scenarios, lightweight heuristics, hierarchical optimization, or tighter integration of prediction and scheduling may be required to preserve responsiveness without compromising quality.

## 7 Related Works

**Model-serving.** Several open source and commercial frameworks like TensorFlow Serving [18], Tensor RT [17], SageMaker [16], Triton [19], and Azure ML [11] exist for ML models. However, they do not readily support scaling for generative models. Recent works [27, 32, 53, 61, 69, 73, 78, 83, 85] also proposed optimized inference serving pipelines for discriminative models, but extending them to generative models like T2I is fairly non-trivial. Also, LLM serving works [74, 75] focusing on the decoding phase [22, 40] and memory management [51, 56, 88] are irrelevant for T2I serving.

**Scaling.** Previous works [23, 37, 65, 85] use VM-level horizontal scaling and model-level autoscaling through model selection, switching, and routing to handle load and optimize accuracy or cost. These are orthogonal to Argus and can be used to increase model instances when the load exceeds model-scaling limits.

**Model optimizations.** Approximation techniques such as distillation [60, 66], pruning [35], sparsification [45, 87], quantization [55, 79, 80], and others [21, 36, 46, 52, 58, 59, 81] accelerate inference while preserving utility. These can be used within AC or SM, or as new approximation strategies, complementing Argus.

**Generative Paradigms in T2I.** Diffusion models underpin most state-of-the-art and production T2I systems, offering superior fidelity, flexible conditioning, and scalability [10, 12–14, 33, 34, 55, 62, 63, 77]. While recent autoregressive models like LlamaGen [70] and Janus Pro [31] show promise, they remain nascent. Emerging autoregressive approaches [54] replace discrete tokenization with per-token diffusion processes, enabling continuous-valued generation while maintaining both speed and quality.

## 8 Conclusion

We presented Argus, a high-throughput inference serving system for text-to-image models that sustains generation quality under high load. Argus employs a novel load assignment algorithm to optimally route prompts across model variants with different approximation levels. By combining approximate caching with selective variant execution, Argus adapts to workload dynamics without incurring model-switching overheads, substantially reducing latency SLO violations while preserving output fidelity.